# Shape-aware Generative Adversarial Networks for Attribute Transfer

Lei Luo[a], William Hsu[a], Shangxian Wang[b]
[a]Kansas State University, [b]Johns Hopkins University

## ABSTRACT

Generative adversarial networks (GANs) have been successfully applied to transfer visual attributes in many domains, including that of human face images. This success is partly attributable to the facts that human faces have similar shapes and the positions of eyes, noses, and mouths are fixed among different people. Attribute transfer is more challenging when the source and target domain share different shapes. In this paper, we introduce a shape-aware GAN model that is able to preserve shape when transferring attributes, and propose its application to some real-world domains. Compared to other state-of-art GANs-based image-to-image translation models, the model we propose is able to generate more visually appealing results while maintaining the quality of results from transfer learning.

## 1. INTRODUCTION

Attribute transfer in vision refers to transferring some abstract elements of source images to images in a target domain. For example, in human face attribute transfer, the task often refers to transforming smiling to neutral faces, while the identity of each face is maintained. Many approaches have been proposed for this task. Since many domains entail limited access to paired data (that is the same person with different facial expressions), attribute transfer mainly depends on cycle-consistency loss [1]. Several studies exploited mappings among different domains under the constraint of cycle-consistency and achieved satisfying results [2]-[3]. Fig. 1 shows that StarGAN and ELEGANT are able to transfer human face attributes, however, their approach does not provide intermediate states from the source to target domain.

Another branch of research studies latent space interpolation, which tries to interpolate between two domains and thus generate a sequence of intermediate states besides just the one target state. This depends on the assumption of attribute space being flat and linear. Convolutional Neural Networks (ConvNets) have achieved great success in image classification in recent years [4]-[7]. The last layer of ConvNets is usually a fully connected layer without activation functions, which linearly map learned features to class labels. Since the features are linearly separable, one can transfer features in a simple fashion. For example, let $x$ and $y$ be learned features of two instances from two different domains. The transferred $x$ can be obtained by moving it towards the direction of $y$. Intermediate states can also be obtained along the transition.

Although some successful attribute transfer has been demonstrated through cycle-consistency loss and latent space interpolation, it remains a challenge to transfer attributes while successfully keeping the identity of source domain after transferring. Taking attribute transfer between human faces as an example, many GANs-based models are capable of adding glasses to faces pictures that do not wear glasses in the first place, or make non-smiling faces into smiling ones. This success partially depends on the fact that the shape of human faces is similar and the positions of face key points, such as the nose and eyes, are relatively fixed. Moreover, data sets such as CelebA, are usually preprocessed in a way that key point positions are aligned in order to generate natural-looking results after transferring. In domains without this feature or among domains that are vastly different, the resulting transferred images can be artificially fake and unnatural-looking to humans if naively applying the techniques mentioned above.

Thus, in this study, we use an example domain of tomato leaf data set, which contains healthy leaves and leaves with different kinds of diseases. We treat different types of leaves as different domains and the goal is to transfer exemplars of healthy ones to those of known categories of unhealthy ones while maintaining their identity, which we deem to be the shape (especially outer contour) of leaves. This is the novel contribution of this study, which aims to generalize to other domains with varying shapes. Our novel contributions are as follows.

- We proposed a novel shape-aware GANs model that is capable of multi-domain, multi-modal attribute transfer while maintaining the shape of source domain.

- We propose several strategies for stabilizing training of the proposed model.
- Experiments showed that our model is able to generate more visually satisfying results than recently proposed state-of-art baseline model while maintaining the quality of translated results.

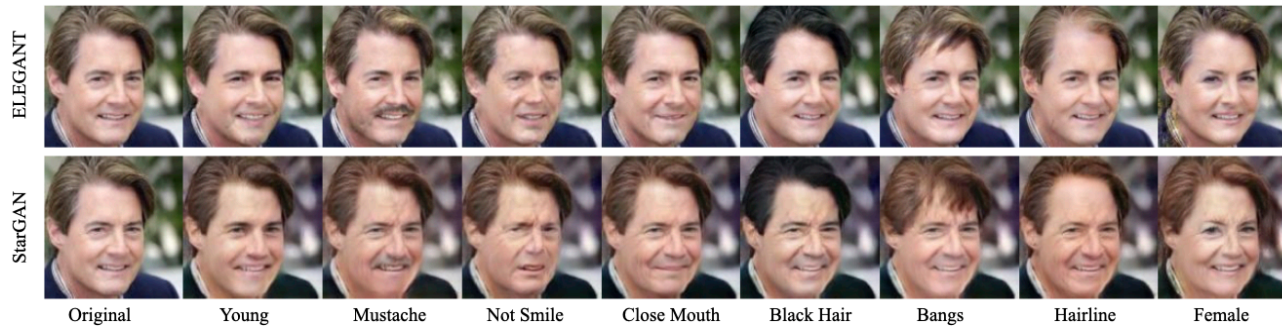

Fig. 1 Examples of transferring human face attributes by StarGAN and ELEGANT. Figure is excerpted from HomoInterpGAN [11]

## 2. RELATED WORK

### 2.1 Image-to-Image Translation

Impressive results have been achieved in recent years in image-to-image translation. Pix2pix [8] produced high quality results. It features conditional GANs and uses adversarial loss and L1 loss to guide the learning of the model. To increase the diversity of produced images, noise term was added to improved version of pix2pix model. Since L1 loss depends on paired data that is difficult to obtain, studies have also focused on unpaired image-to-image translation, where CycleGAN [1] and DiscoGAN [9] were later developed. They utilize the cycle-consistency loss to map between two domains while maintaining some key attributes. However, they are mostly able to train two domains at a time and often applied in facial expression field. Moreover, naively applying cycle-consistency loss does not guarantee production of natural-looking results when domains are highly disparate from one another and internally diverse.

### 2.2 Latent Space Interpolation

One drawback of GANs that solely depends on cycle-consistency loss is an inability to produce sequence of intermediate images from the source to target domain. Latent space interpolation builds on the fact that there is a flat feature space [10]. Once the original image space is mapped onto a feature space, interpolation can be done by gradually moving the latent space of the source domain towards the target domain. However, there are infinitely many ways of connecting two points in the latent space, and thus finding the one that can produce smooth and natural-looking results is of great value. Instead of naively interpolating using a straight line connecting the two points in latent space, [11] uses an artificial neural network (ANN) to learn the path and achieves visually satisfying results.

### 2.3 Image Segmentation

One of first deep learning models for image segmentation was proposed using a fully convolutional network (FCN) [12], which only has convolutional layers and can produce a segmentation maps of same size of input images. One of the drawbacks of FCN is that it pays little attention to useful scene-level semantic context. To remedy this problem, deep learning model incorporated with graphical models were proposed. Several learning frameworks incorporate Conditional Random Fields [13]-[15]. Markov Random Fields are representations that are also commonly used to apply deep learning models for image segmentation [16]. Another branch of work for image segmentation uses an encoder-decoder architecture. Some early work was done by [17], where the encoder uses similar architecture as VGG-16 and the decoder consists of deconvolution and unpooling layers. The UNet [18], initially developed for medical image segmentation, is also commonly used. A UNet consists of down-sampling and up-sampling steps, where the former extracts features by using $3 \times 3$ convolutions and the latter reduces the number of feature maps while increasing the dimension. Features from down-sampling are concatenated with those from up-sampling and finally $1 \times 1$ convolution is used to generate segmentation map.

## 3. METHOD

Without loss of generality, we take transferring attributes from two different domains as an example to illustrate the proposed model. We would like to transfer attribute from domain $X$ to image example of domain $Y$.

### 3.1 Learning Encoder and Decoder

Images $x$ are passed into an encoder ($E$) first, resulting latent vector ($V$) of fixed length, thus $V_x = E(x)$. The image from target domain $Y$ also goes through the encoder and obtain $V_y$. Next, an ANN (ann) is used to help guide the transition from $V_x$ to $V_y$, and the interpolated latent vector is obtained by $V_I = ann(V_x, V_y)$. Eventually, the decoder ($D$) generates the interpolated image by $D(V_I)$. The loss for the encoder is:

$$L_E = -\mathrm{E}_{x \sim X,\ y \sim Y} \left( ann(E(x), E(y)) \right) \tag{1}$$

In order to guide the learning of the decoder, real images are compared to reconstructed ones and reconstruction loss is formulated as:

$$L_{recons} = MSE\left(x, D(E(x))\right) \tag{2}$$

$$L_{E,ann} = \mathrm{E}_{F \sim P_r}[\mathfrak{D}(F)] - \mathrm{E}_{\tilde{F} \sim P_f}[\mathfrak{D}(\tilde{F})] \tag{3}$$

### 3.2 Learning Critic

Similar to WPGAN [19], the loss function for learning the critic ($\mathfrak{D}$) is formulated as:

$$L_{\mathfrak{D}} = \mathrm{E}_{\tilde{F} \sim P_f}[\mathfrak{D}(\tilde{F})] - \mathrm{E}_{F \sim P_r}[\mathfrak{D}(F)] + \lambda_{gp} GP \tag{4}$$

where $\tilde{F} = ann(F_x, F_y)$ is the interpolated feature, $F_x = E(x)$ and $F_y = E(y)$ are extracted features from the encoder ($E$). $GP$ is the gradient penalty term and $\lambda$ are defined in [19]. $P_f$ and $P_r$ are distributions from fake and real feature samples. $\lambda_{gp}$ is set to 10 for all experiments in this study.

### 3.3 Learning Interpolator

After the encoder projects images into latent space, which is flat, interpolation can be done linearly as:

$$f(F_x, F_y) = F_x + \alpha(F_y - F_x) \tag{5}$$

where $\alpha$ controls the interpolation strength.

Since there are many paths connecting two points in the latent space, naively interpolate linearly might produce blurry images with many artifacts, an ANN is employed to model the best transition from two domains. Our interpolation method is formulated as:

$$f(F_x, F_y) = F_x + \alpha \times ann(F_y - F_x) \tag{6}$$

where $ann$ is a learnable CNN.

### 3.4 Learning Shape

In order to preserve the identity of the source domain, our model incorporates UNet, which outputs a binary map delineating the boundary of leaves, for calculating the shape loss ($L_{shape}$), which is formulated as:

$$L_{shape} = Dice(S_{interpolated}, S_x) \tag{7}$$

where $Dice$ is the Dice Loss [20], $S_{interpolated}$ and $S_x$ are the UNet output of interpolated image and source domain image, respectively.

### 3.5 Model Architecture and Training Algorithm

The proposed model is presented in Figure 2. Input images and target domain images are fed into $E$ first, which produces respective features. The features are then piped into the interpolator, which guides the transition from source to target domain and produces the interpolated feature. Features from target domain and interpolated feature are processed by the critic, which calculates the $L_\mathfrak{D}$ and $L_E$. The decoder maps features from source domain images to the original image space, and then $L_{recons}$ is calculated by comparing the reconstructed image and original image. A UNet is also trained to preserve the shape of interpolated images by calculating the Dice loss ($L_{shape}$). The training procedure is shown in Algorithm 1.

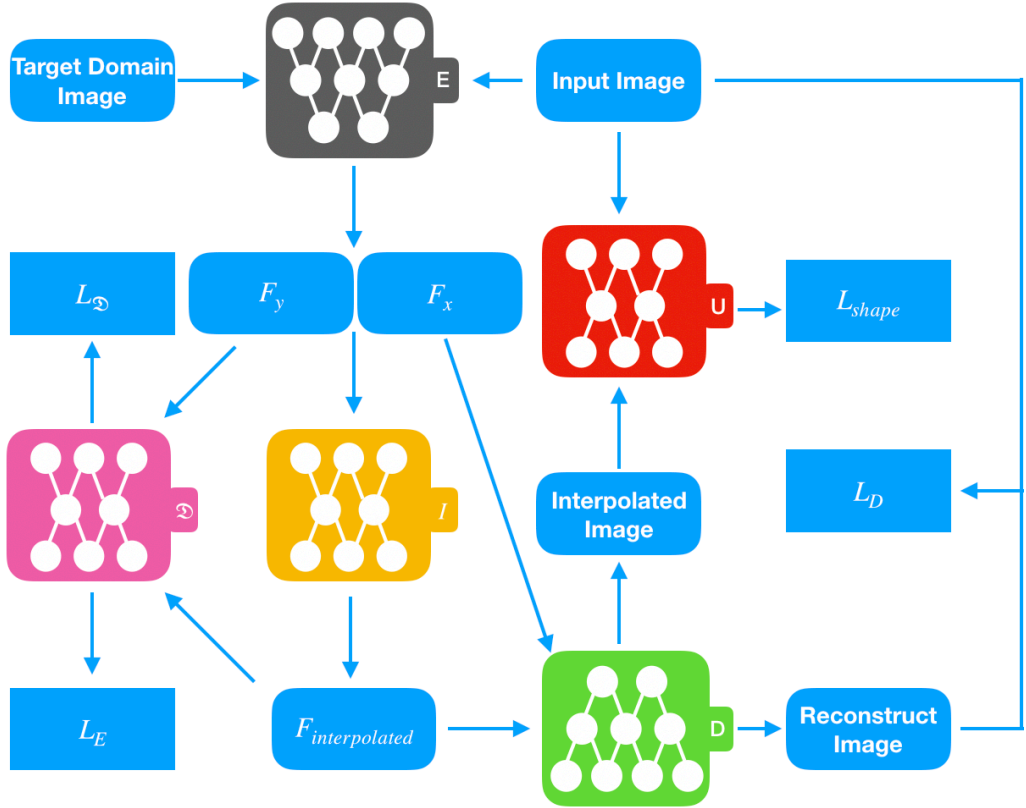

Fig. 2 The structure of our framework. the encoder $E$ maps images of source domain and target domain to their feature space. The interpolator $I$ learn the path from source to target latent space. The decoder $D$ reconstructs the source image from feature space $F_x$ and produces interpolated images from interpolated features. The critic $\mathfrak{D}$ learns how real the interpolated features are. UNet $U$ forces the shape our interpolated images and source images to be similar.

**Algorithm 1**: Model Training, $n_{critic} = 5$
**Data**: Source domain images $x_i$ and target domian images $y_i$, where $i = 1,2,...,N$
**Result**: encoder $E$, decoder $D$, interpolator $I$, and UNet $U$

Initialization;
**while** not converge **do**
    $m = 0$
    **while** $m < n_{critic}$ **do**
        Calculate $L_\mathfrak{D}$ and update $\mathfrak{D}$;
        Calculate $L_{recons}$ and update $D$;
        $m++$;
    Calculate $L_E$ and $L_{E,ann}$ update $E$ and $I$;
    Calculate $L_{shape}$ and update $U$;

## 4. EXPERIMENTS AND RESULTS

We experimented with the PlantVillage tomato leaf data set. It contains healthy leaves and ill leaves, which are categorized into *bacterial spot*, *early blight*, *late blight*, *mold*, *septoria spot*, *spider mites*, *target spot*, and *yellow leaf*. All the images are resized to $128 \times 128$. We annotated the segmentation map of leaves for training the UNet module. The baseline model we compared against is HomoInterpGAN, which achieved impressive results in transferring human-face attributes.

Both our model and the baseline model are capable of interpolating intermediate images from source to target domains. The results are shown in Fig. 3, which illustrates the interpolation results from source (healthy leaf) to three different target domains (three types of unhealthy leaves) under incremental transition strength. Results show that both models are capable of interpolation in latent space and produce intermediate images from source to target domain. However, interpolation results from HomoInterpGAN are not able to preserve the shape of source images. We observe some ghosting effects, where the interpolation tries to copy the shape of target domain in an implausible manner, as shown in row ($a$) with transition strength 0.75 and 1. Another example shown in row ($c$) with transition strength 1, produced by HomoInterpGAN, develops another pointy tip, which seems to be inherited from the target leaf. The incapability of HomoInterpGAN in preserving shape is most obvious in row ($e$) with transition strength 0.75 and 1, where the shape of source image was totally transformed into that of the target domain. Image translation results showed that our model is able to much better preserve the shape of source domain while constraining the results to attribute of the target domain, producing natural-looking interpolation images.

As in [1], [8], [21], we trained a neural network (ResNet34) as a classifier to test the quality of produced image after image-to-image translation. The trained classifier was tested on the original testing data set and the test set accuracy is treated as the baseline. HomoInterpGAN and our model were used to transform healthy leaves into unhealthy ones, and the produced images are then classified. Table 1 shows the test set accuracy of the classifier trained on the original data set and that of produced images after image-to-image translation. Compared to the baselines testing accuracy (about 97%), about 95% translated images by HomoInterpGAN are correctly classified into corresponding categories and nearly 94% of translated images by our model are successfully identified. Albeit being slightly lower than HomoInterpGAN on test accuracy, images produced by our model are more visually satisfying, that is to say, qualitative attributes from target domain are successfully transferred to the source domain and the shape of source images is preserved.

TABLE I
TEST ACCURACY ON DIFFERENT DATA SETS

| Data set | Accuracy |
| --- | --- |
| Original testing data set (baseline) | 97.33% |
| Translated images by HomoInterpGAN | 95.02% |
| Translated images by ours | 93.87% |

## 5. CONCLUSIONS

We have established a new framework for unpaired image-to-image translation, transferring attributes, and producing natural-looking intermediate results. Our model features a UNet module that preserves the shape of translation results. In addition, our model incorporates a neural network that is especially designed for learning the best path to transit in the latent space. Results showed that both our model and the baseline model are able to transfer attributes. However, in transforming images, the baseline model naively attempts to copy the shape of target domains, and thus generates ghosting artifacts in the translated results. By contrast, our approach is capable of smoothly transferring attributes and producing more visually appealing results by preserving the shape of source domain without too much trade-off (about 1%) on the quality of translated results.

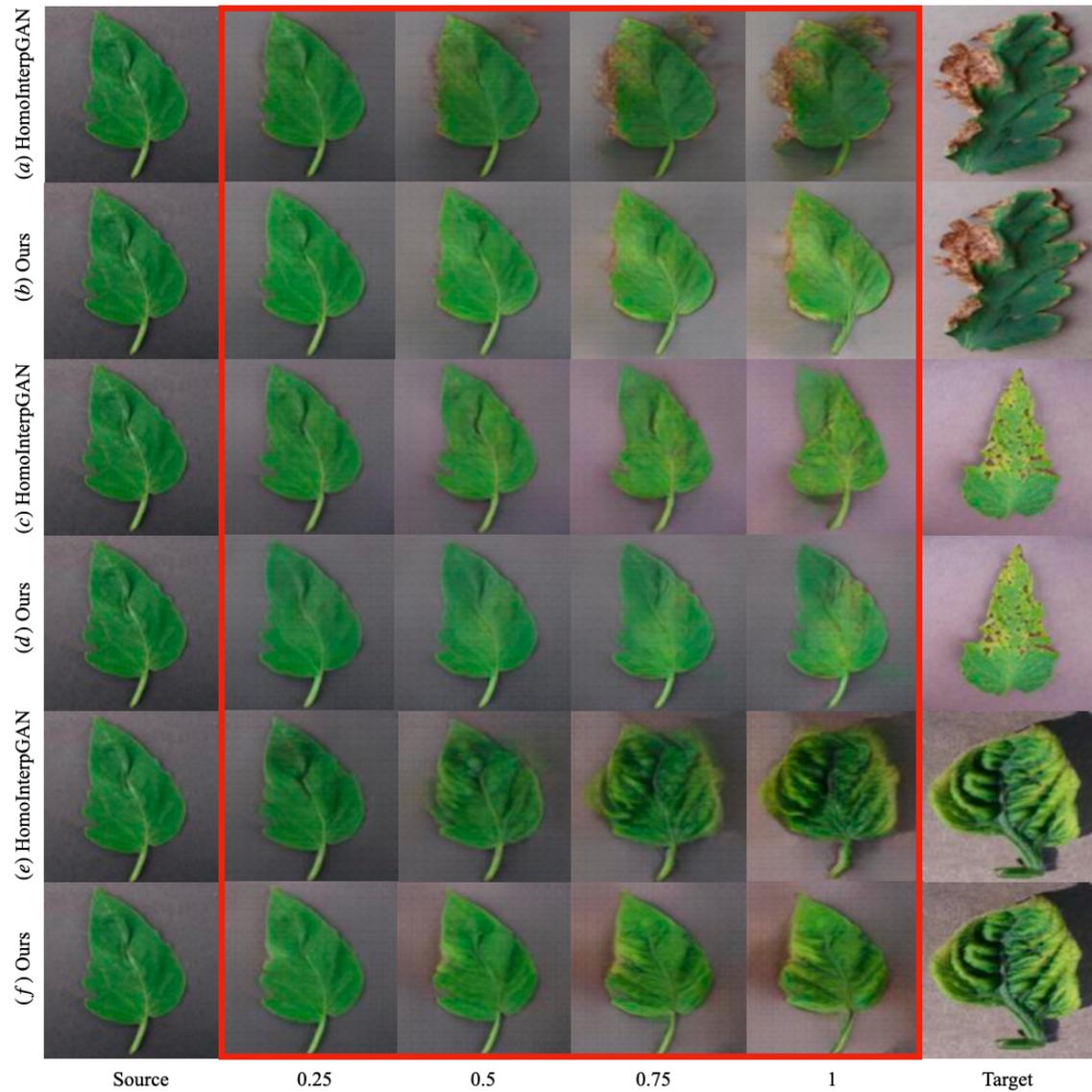

Fig. 3 Interpolation results from our model and HomoInterpGAN. The source image is one of the healthy leaves. The target of (*a*) and (*b*) is one example from bacterial leaves. The target in (*c*) and (*d*) is one of the septoria leaves. The target of (*e*) and (*f*) is a yellow leaf. Images in red box are interpolation results, where 0.25, 0.5, 0.75, and 1 represent the interpolation strength.